\documentclass[draftclasofoot, onecolumn]{IEEEtran}
\IEEEoverridecommandlockouts
\usepackage{array}
\usepackage[caption=false]{subfig}
\usepackage[noadjust]{cite}
\usepackage[bookmarks=true,breaklinks=true,colorlinks,linkcolor=red,citecolor=blue,urlcolor=BlueViolet]{hyperref}
\usepackage{multirow}
\usepackage{amsmath,amssymb,amsfonts}

\usepackage{graphicx}
\usepackage{threeparttable}
\usepackage{textcomp}
\usepackage{adjustbox} 
\usepackage{xcolor}
\def\BibTeX{{\rm B\kern-.05em{\sc i\kern-.025em b}\kern-.08em
    T\kern-.1667em\lower.7ex\hbox{E}\kern-.125emX}}
\usepackage{booktabs}
\usepackage{tablefootnote}
\usepackage{float}
\usepackage{algorithmic}
\usepackage{textcomp}
\usepackage{booktabs}

\usepackage{makecell}
\usepackage{color}
\usepackage{soul}
\usepackage{mdframed}
\renewcommand{\hl}[1]{#1}

\renewcommand{\figurename}{Figure}
\renewcommand{\thetable}{\arabic{table}}
\renewcommand{\tablename}{Table}

\begin{document}

\title{Spectra-Guided Neural Tucker Factorization}

\author{Fusheng Wang \quad Yikai Hou\textsuperscript{*}\thanks{Fusheng Wang is with the School of Automation, Chongqing University of Posts and Telecommunications, Chongqing, China (fushengw@outlook.com). Yikai Hou is with the College of Computer and Information Science, School of Software, Southwest University, Chongqing, China (yikaih@email.swu.edu.cn). \textsuperscript{*}Corresponding author. This research is supported by the National Key Research and Development Program of China under grant 2024YFF0908200, the Fundamental Research Funds for the Central Universities SWU-KR24005, the Southwest University Graduate Research Innovation Project under grant SWUB25054, and Chongqing Natural Science Foundation under grants CSTB2024TIAD-CYKJCXX0028 and CSTB2025TIAD-STX0032.}}

\maketitle

\begin{abstract}
This paper proposes Spectra-Guided Neural Tucker Factorization (SG-NTF) for High-Dimensional 
and Incomplete (HDI) tensor completion. Circumventing discrete representational limits, SG-NTF 
maps scalar timestamps into a continuous spectral space to abstract temporal periodicities. 
Concurrently, a Spatio-Temporal Co-Gating (STCG) mechanism explicitly filters latent interactions 
via multiplicative modulation on spatiotemporal contexts. Evaluations on real-world HDI tensors 
verify that SG-NTF maintains competitive completion accuracy with parameter efficiency.
\end{abstract}


\section{Introduction}

Multidimensional observations from traffic monitoring \cite{ChenX22} and Quality-of-Service 
(QoS) tracking \cite{TangM16, LuoX19_} exhibit complex spatiotemporal 
dependencies. Due to partial observability, they are naturally represented as High-Dimensional 
and Incomplete (HDI) tensors \cite{WangQ22}. 
Consequently, Latent Factorization of Tensors (LFT) via Tucker decomposition \cite{TG09} 
offers an expressive framework for missing data estimation \cite{ZhangW14, YeF21} 
by extracting a core tensor to explicitly capture high-order latent interactions \cite{HouY26,TangP26_,HouY25,ZhangX24,MiJ23,XuX23,LinM25,GC25,BZ24,ChengS24,ChenH25,Lu25,ChenH24,LinM25_,YangH25,WuD23,WuD25,ZhenY22,LuoX21_,LuoX23_,TangP24,ChenM24,PengZ22,ChenM25,SuX21,HeQ19,LuoX14,LuoX19__,LuoX14__,LuoX18_,LiuZ18,WuH22,WuH21_,WuH21__,WuH24,TangP24_,LuoX20,WuH22__,HeX17}.

Recent neural tensor factorization models \cite{HeX17, TangP24_} indicate that applying 
nonlinear filtering to the core tensor, namely \textit{core zeroization}, isolates task-relevant 
components to improve representation accuracy \cite{LuoX23_}. While deep auto-encoding 
modules successfully approximate this filtering \cite{TangP24_}, achieving it with 
reduced parameter complexity remains highly desirable \cite{BiF22,WuH22+,LuoX21__,LiW23,LiW22+,ChenM21+,ZhongY23,LuoX21____,ZhouY21,LiuZ20,LiuZ21,ChenJ21,LiM21+,WuD22,LuoX21,LuoX16,LuoX18,LuoX22_,ChenJ23,LuoX21+,LuoX22+,LiJ24,ZhongY24,YuanY24,LiW22,LuoX21++,WuH22_,ShiX20,YuanY20,WuD21+,WuD21++,WangJ24+,WuD23+,YuanY24+,YuanY23+,WuH21++,ChenM24++,WuH24+,WangQ22+,HuL21+,YanJ23,QiY21,HuL21,HeY21,SongY22,ChenH24_,XuX25,LuoX22__,HeX20,A17,WuD21,ChenD21}. Concurrently, standard discrete temporal 
embeddings restrict the capacity to abstract continuous periodicities.

This paper proposes Spectra-Guided Neural Tucker Factorization (SG-NTF). Dictated by Bochner’s 
theorem \cite{RahimiR07}, an integrated Learnable Fourier Feature Mapping (LFFM) module 
projects timestamps into a continuous spectral space to capture periodic fluctuations 
\cite{TancikM20, A17}. Concurrently, a Spatio-Temporal Co-Gating (STCG) 
mechanism executes context-aware feature modulation \cite{SrivastavaR15}.
By calibrating latent interactions via a Hadamard product, STCG accommodates location-specific 
spatial dynamics, serving as a parameter-efficient alternative to deep auto-encoders. Empirical 
studies demonstrate that SG-NTF maintains competitive completion performance with significantly 
reduced complexity.

\section{Preliminaries}

\subsection{Tensor Notation and HDI data}

Let $\mathcal{Y} \in \mathbb{R}^{I \times J \times K} $ denote a three-way tensor 
with known $\Lambda$ and unknown $\Gamma$ entries. It is 
formally categorized as an HDI tensor if $|\Lambda| \ll |\Gamma|$\cite{TangP24_}. 
Tensor completion infers missing entries in $\Gamma$ by optimizing a low-rank 
approximation $\hat{\mathcal{Y}}$ exclusively over $\Lambda$\cite{TG09}.

\subsection{Neural Tucker Factorization}

NeuTucF \cite{TangP24_} casts classic Tucker decomposition \cite{TG09} into a neural framework.
 For an observation $y_{ijk} \in \Lambda$, mode indices are parameterized into latent embeddings $\mathbf{a}_i \in \mathbb{R}^P$, $\mathbf{b}_j \in \mathbb{R}^Q$, 
and $\mathbf{c}_k \in \mathbb{R}^R$. The interaction tensor is vectorized into  
$\mathbf{t}_{ijk} = \mathcal{F}(\mathbf{a}_i \circ \mathbf{b}_j \circ \mathbf{c}_k) \in \mathbb{R}^{P \times Q \times R}$, 
projecting into a scalar estimation:
\begin{equation}
    \hat{y}_{ijk} = \sigma \big( \mathbf{w}_{core}^\top \boldsymbol{t}_{ijk} \big),
\end{equation}
where $\mathbf{w}_{core} \in \mathbb{R}^M$ conceptually functions as the trainable core tensor, and 
$\sigma(\cdot)$ denotes the logistic sigmoid activation. 

\section{Spectra-Guided Neural Tucker Factorization}
\begin{figure}
    \centering
    \includegraphics[width=\linewidth]{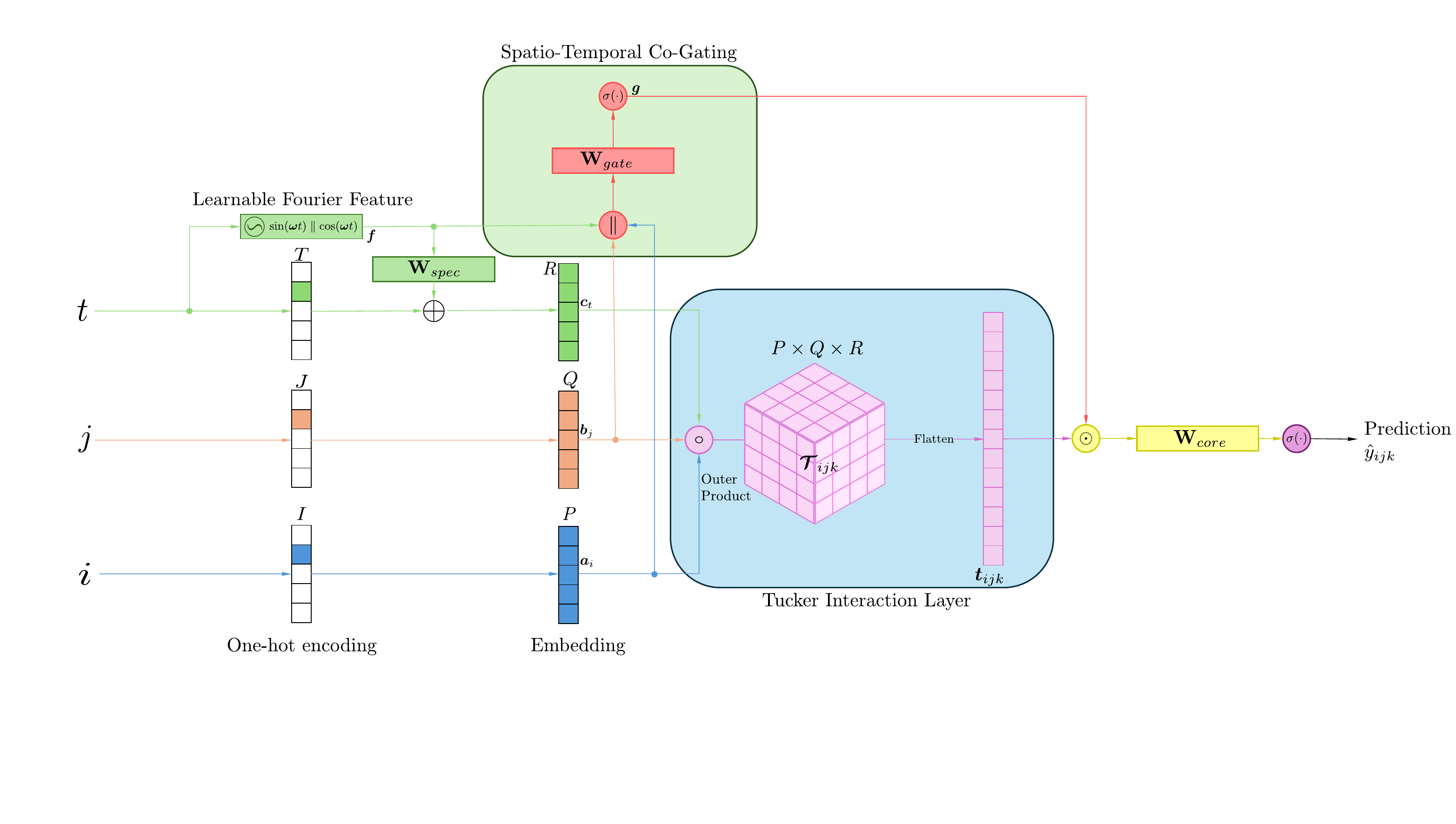}
    \caption{Spectra-Guided Neural Tucker Factorization (SG-NTF) model.}
    \label{fig:framework}
\end{figure}

Figure \ref{fig:framework} depicts SG-NTF. Deviating from discrete representations, it parameterizes 
the temporal mode via a continuous spectra-guided embedding to abstract functional periodicities. 
Following an outer product among mode embeddings, a lightweight Spatio-Temporal Co-Gating (STCG) 
mechanism dynamically modulates the raw latent interactions.

\subsection{Spectra-Guided Continuous Temporal Embedding}

Conventional LFT models parameterize the temporal mode via independent latent features, failing to 
accurately capture continuous temporal dependencies in HDI data. To address this, a continuous 
functional representation is constructed. Dictated by Bochner’s theorem \cite{RahimiR07}, 
stationary temporal patterns are rigorously abstracted by a continuous frequency distribution:
\begin{equation}
    \kappa(t, t') = \kappa(t - t') = \int_{\mathbb{R}} e^{i\omega(t-t')} d\mu(\omega).
\end{equation}
It indicates that complex dependencies can be effectively approximated via Fourier basis functions.

Guided by this, a Learnable Fourier Feature Mapping (LFFM) module \cite{TancikM20} is 
incorporated. 
It maps the \hl{raw discrete timestamp $t \in \{0, 1, \dots, T-1\}$} into 
a spectral space via learnable angular bases $\boldsymbol{\omega } \in \mathbb{R}^{d_{spec}}$, 
($d_{spec}$ denotes spectral dimensionality). \hl{Explicit normalization of $t$ is bypassed, 
as $\boldsymbol{\omega}$ mathematically absorbs scalar normalization factors during optimization:}
\begin{equation}
    \mathbf{f}(t) =[\sin(\boldsymbol{\omega}t) \parallel \cos(\boldsymbol{\omega}t)] \in \mathbb{R}^{2d_{spec}},
\end{equation}
where $\parallel$ denotes concatenation. Unlike fixed positional encodings, enforcing the 
learnability of $\boldsymbol{\omega}$ enables dynamic abstraction of latent periodicities presented in HDI tensors.

\hl{The learnable spectral bases $\boldsymbol{\omega}$ are initialized as an 
evenly spaced grid over $[\frac{2\pi}{T}, \pi]$. The lower bound $\frac{2\pi}{T}$ 
corresponds to the maximum observable temporal span $T$ to capture long-term trends, 
whereas the upper limit $\pi$ aligns with the Nyquist frequency to prevent aliasing.}

While Bochner's theorem theoretically abstracts stationary periodicities, 
real-world temporal dynamics exhibit non-stationary shifts resisting strictly periodic 
approximations. To accommodate such irregularities, a discrete residual embedding $\mathbf{e}_{res} \in \mathbb{R}^{R}$ 
augments the continuous spectral representation, specifically absorbing 
localized temporal deviations:
\begin{equation}
    \mathbf{c}_t = \mathbf{W}_{spec} \mathbf{f}(t) + \mathbf{e}_{res}(t),
\end{equation}
where $\mathbf{W}_{spec} \in \mathbb{R}^{R \times 2d_{spec}}$ is a  
linear projection matrix. This formulation preserves functional interpolation capabilities 
alongside representational flexibility for localized irregularities.

Regarding the convergence of the learnable angular frequencies
 $\boldsymbol{\omega}$, the applied sinusoidal basis functions and their partial derivatives remain 
 rigorously bounded over a finite temporal domain. Composed with the 
 Lipschitz continuous downstream logistic sigmoid activation, the overall optimization 
 objective maintains $L$-smoothness concerning $\boldsymbol{\omega}$. Consequently, under 
 standard stochastic gradient descent, parameter updates converge to a stationary point \cite{GhadimiL13, BottouC18}.

The continuous embedding $\mathbf{c}_t$ subsequently undergoes an outer product alongside spatial 
embeddings $\mathbf{a}_i \in \mathbb{R}^P$ and $\mathbf{b}_j \in \mathbb{R}^Q$, yielding the raw interaction tensor:
\begin{equation}
    \boldsymbol{\mathcal{T}}_{ijt} = \mathbf{a}_i \circ \mathbf{b}_j \circ \mathbf{c}_t.
\end{equation}
However, this standard outer product imposes an implicit homogeneity constraint  
across spatial entities. Such rigid mapping restricts the representational capacity 
to characterize location-specific fluctuations. To circumvent this, a context-aware 
gating mechanism is devised.

\subsection{Spatio-Temporal Co-Gating}

Imposing global temporal periodicities uniformly across spatial entities ignores 
context-dependent spatial dynamics. 
Latent temporal patterns are entangled with specific spatial contexts, as
distinct spatial entities exhibit heterogeneous responses to the same periodicity due to topological disparities.  
Consequently, a uniform mapping decouples spatial and temporal modes, failing to accurately capture complex nonlinear 
spatiotemporal dependencies.

To address this, a Spatio-Temporal Co-Gating (STCG) mechanism explicitly generates a 
dynamic filter from the concatenated spatiotemporal context. The raw interaction 
tensor $ \boldsymbol{\mathcal{T}_{ijt}} =\mathbf{a}_i \circ \mathbf{b}_j \circ \mathbf{c}_t $ 
is vectorized into  $\mathbf{t}_{ijt} = \mathcal{F}(\boldsymbol{\mathcal{T}}_{ijt}) \in \mathbb{R}^{P \times Q \times R}$.

Specifically, the STCG concatenates the Fourier features $\mathbf{f}(t)$ with spatial mode embeddings $\mathbf{a}_i$ and $\mathbf{b}_j$ to compute a personalized dynamic gate $\mathbf{g}(t,i,j) \in (0,1)^M$:
\begin{equation}
    \mathbf{g}(t,i,j) = \sigma \Big( \mathbf{W}_{gate} \big[ \mathbf{f}(t) \parallel \mathbf{a}_i \parallel \mathbf{b}_j \big] \Big),
\end{equation}
where $\mathbf{W}_{gate} \in \mathbb{R}^{M \times (2d_{spec} + P + Q)}$ denotes the projection weight matrix and $\sigma(\cdot)$ is the sigmoid activation function. 
\hl{To address scale disparity between bounded Fourier features and spatial embeddings, 
$\mathbf{W}_{gate}$ adaptively rescales the concatenated inputs, ensuring balanced spatiotemporal 
contributions during optimization.}

This dynamic gate applies element-wise multiplicative modulation on the flattened interaction 
vector, enforcing context-aware core zeroization: 
\begin{equation}
    \tilde{\mathbf{t}}_{ijt} = \mathbf{t}_{ijt} \odot \mathbf{g}(t,i,j),
\end{equation}
where $\odot$ denotes the Hadamard product.
\hl{The gated features $\tilde{\mathbf{t}}_{ijt}$ represent localized 
spatiotemporal patterns. While the raw interaction vector encodes global 
periodicities across all spatial topologies, the dynamic gate operates as 
a location-specific filter. By attenuating irrelevant interaction channels, 
the gated features isolate the specific physical dynamics governing each observation.}


\subsection{Output Mapping and Complexity Analysis}
A linear projection layer maps the modulated interaction vector $\tilde{\mathbf{t}}_{ijt}$ 
into a scalar approximation:
\begin{equation}
    \hat{y}_{ijt} = \sigma \big(\mathbf{w}_{core}^\top \tilde{\mathbf{t}}_{ijt} \big),
\end{equation}
where $\mathbf{w}_{core} \in \mathbb{R}^M$ conceptually resembles the vectorized 
Tucker core tensor. \hl{Since HDI tensors are normalized into $(0, 1)$, observations 
represent probabilistically bounded variables. Because linear projections yield 
unconstrained real values, the logistic sigmoid activation functions as an inverse 
link function, mapping unbounded projections into the $(0,1)$ interval to align with this 
probabilistic assumption.}

Despite dynamic modulation, the architecture incurs negligible parameter overhead compared with 
the NeuTucF baseline. Operating as a soft attention mechanism, STCG bypasses deep non-linear 
layers to preserve structural transparency. It explicitly quantifies how spatial entities 
attenuate or amplify latent interactions in response to temporal periodicities. Furthermore, 
while preserving the canonical Euclidean objective, parameterizing interactions via spatiotemporal 
contexts constrains the hypothesis space. This formulation serves as implicit regularization, 
shifting optimization dynamics to learn a context-aware filter.

\section{Experiments}

\subsection{General Setting}

\textit{Datasets:} This work evaluates the proposed Spectra-Guided Neural Tucker 
Factorization (SG-NTF) model on six HDI tensor datasets (denoted as D1-D6). 
Extracted from three publicly accessible datasets (NYCTaxi, PCTemp, and WSDream), they 
exhibit known data densities from 1.25\% to 19.36\%. These evaluated datasets are summarized in Table \ref{tab:dataset_details}. 
As their raw data distributions are highly 
skewed with large variances, data preprocessing via log transformation and min-max normalization 
into $(0,1)$ is performed in all experiments to fit the probabilistic assumption of 
low-rank factorization. 

\begin{table}
\centering
\caption{Details of Experiment Datasets.}
\label{tab:dataset_details}
\setlength{\tabcolsep}{1.5mm}{
\begin{tabular}{clcccc}
\toprule
\textbf{No.} & \textbf{Dataset} & \textbf{Dimensions ($I \times J \times T$)} & \textbf{\makecell{Split\\Ratio}} & \textbf{\makecell{Density\\(\%)}} & \textbf{\makecell{Sparsity\\(\%)}} \\
\midrule
\textbf{D1} & \multirow{2}{*}{NYCTaxi} & \multirow{2}{*}{$30\times30\times1464$} & 10:90 & 7.346 & 92.654 \\
\textbf{D2} & & & 20:80 & 14.692 & 85.308 \\
\midrule
\textbf{D3} & \multirow{2}{*}{PCTemp} & \multirow{2}{*}{$30\times84\times399$} & 10:90 & 9.678 & 90.322 \\
\textbf{D4} & & & 20:80 & 19.356 & 80.644 \\
\midrule
\textbf{D5} & WSDream (QoS-RT) & \multirow{2}{*}{$142\times4500\times64$} & 2:98 & 1.329 & 98.671 \\
\textbf{D6} & WSDream (QoS-TH) & & 2:98 & 1.255 & 98.745 \\
\bottomrule
\end{tabular}}
\end{table}

\textit{Compared Models:} The proposed SG-NTF model is compared with three models for 
missing entry prediction in an HDI tensor. The compared models include: 1) M1: A CP-based tensor 
factorization model; 2) M2: An ensemble neural network model derived from recent MLP and CP 
frameworks; 3) M3: The foundational neural Tucker factorization (NeuTucF) model. All experiments 
are conducted on a workstation equipped with an Intel Core i7-14700F CPU, an NVIDIA RTX 5060 GPU, 
and 16GB of RAM, operating on Windows Subsystem for Linux (WSL) with Ubuntu 22.04.

\textit{Evaluation Metrics:} To evaluate the model performance on predicting unknown 
element entries of the HDI tensor, this paper adopts Mean Absolute Error (MAE), 
Mean Relative Error (MRE), and Root Mean Square Error (RMSE) as the evaluation metrics.

\subsection{Performance Comparison}

The latent rank is uniformly set to $P = Q =R = 5$, and the spectral 
dimensionality is fixed at $d_{spec} = 16$.
Each model is executed 
10 times to calculate mean metrics. The experimental results 
in Table \ref{table_optimized_results} yield the following insights: 

1) SG-NTF is an effective low-rank tensor completion model. As illustrated in Table \ref{table_optimized_results}, 
SG-NTF outperforms M1, M2, and M3 across all three types of metrics. Notably, it achieves maximum MAE accuracy 
gains of 13.56\% and 33.36\% over the foundational baseline M3 on D1 and D3.

2) For accurate representation learning on HDI tensor, it is essential 
to capture continuous temporal periodicities alongside context-dependent 
spatial dynamics. The results on MRE show that SG-NTF can learn much more 
accurate representations compared with other models.

\begin{table}[t]
\centering
\caption{The Summary of Results.}
\label{table_optimized_results}
\setlength{\tabcolsep}{1.5mm}{
\begin{tabular}{cllcccc} 
\toprule
\textbf{No.} & \textbf{Dataset} & \textbf{Metric} & \textbf{M1} & \textbf{M2} & \textbf{M3} & \textbf{SG-NTF} \\
\midrule
\multirow{3}{*}{D1} & \multirow{3}{*}{\makecell{NYCTaxi \\ (10\%)}} 
    & MAE & 5.7569 & 4.1827 & 3.9759 & \textbf{3.4367} \\
  & & MRE & 0.8550 & 0.5552 & 0.5068 & \textbf{0.4532} \\
  & & RMSE & 10.0877 & 7.3428 & 7.1218 & \textbf{5.9100} \\
\midrule
\multirow{3}{*}{D2} & \multirow{3}{*}{\makecell{NYCTaxi \\ (20\%)}} 
    & MAE & 4.4902 & 3.5514 & 3.8535 & \textbf{3.1962} \\
  & & MRE & 0.6080 & 0.4678 & 0.4895 & \textbf{0.4230} \\
  & & RMSE & 8.1637 & 6.2053 & 6.9337 & \textbf{5.4574} \\
\midrule
\multirow{3}{*}{D3} & \multirow{3}{*}{\makecell{PCTemp \\ (10\%)}} 
    & MAE & 2.2144 & 1.0988 & 0.8575 & \textbf{0.5714} \\
  & & MRE & 0.0910 & 0.0446 & 0.0349 & \textbf{0.0229} \\
  & & RMSE & 2.7387 & 1.4103 & 1.1429 & \textbf{0.7491} \\
\midrule
\multirow{3}{*}{D4} & \multirow{3}{*}{\makecell{PCTemp \\ (20\%)}} 
    & MAE & 1.6802 & 0.8092 & 0.8555 & \textbf{0.4744} \\
  & & MRE & 0.0686 & 0.0325 & 0.0349 & \textbf{0.0189} \\
  & & RMSE & 2.1474 & 1.0765 & 1.1431 & \textbf{0.6187} \\
\midrule
\multirow{3}{*}{D5} & \multirow{3}{*}{\makecell{QoS-RT \\ (2\%)}} 
    & MAE & 0.8208 & 0.7434 & 0.6804 & \textbf{0.6487} \\
  & & MRE & 1.9261 & 1.8144 & 1.3210 & \textbf{1.2247} \\
  & & RMSE & 2.2193 & 1.9726 & 1.8770 & \textbf{1.7983} \\
\midrule
\multirow{3}{*}{D6} & \multirow{3}{*}{\makecell{QoS-TH \\ (2\%)}} 
    & MAE & 5.9336 & 5.8545 & 4.8799 & \textbf{4.3689} \\
  & & MRE & 1.0400 & 1.1817 & 0.9159 & \textbf{0.8549} \\
  & & RMSE & 39.2872 & 38.6867 & 35.6002 & \textbf{32.8818} \\
\bottomrule
\end{tabular}}
\end{table}

\subsection{Ablation Study}

Ablation studies on the PCTemp (10:90) dataset decouple the SG-NTF framework into 
two degraded variants:

1) SG-NTF-\textit{w/o}-LFF: Parameterizes the temporal mode via discrete embeddings. 

2) SG-NTF-\textit{w/o}-SC: Ablating spatial context (SC) inputs degenerates STCG to rely exclusively on global Fourier features 
$\mathbf{f}(t)$, bypassing spatial embeddings $\mathbf{a}_i$ and $\mathbf{b}_j$.

Table \ref{tab:ablation} details the performance degradation of these variants, yielding two analytical insights. Firstly, reverting to discrete embeddings (SG-NTF-\textit{w/o}-LFF) severely restricts representational capacity. 
Secondly, bypassing spatial embeddings (SG-NTF-\textit{w/o}-SC) confirms that latent 
temporal dynamics are entangled with spatial topologies.

\subsection{Impact of Spectral Dimensionality}
\begin{figure}
    \centering
    \includegraphics[width=\linewidth]{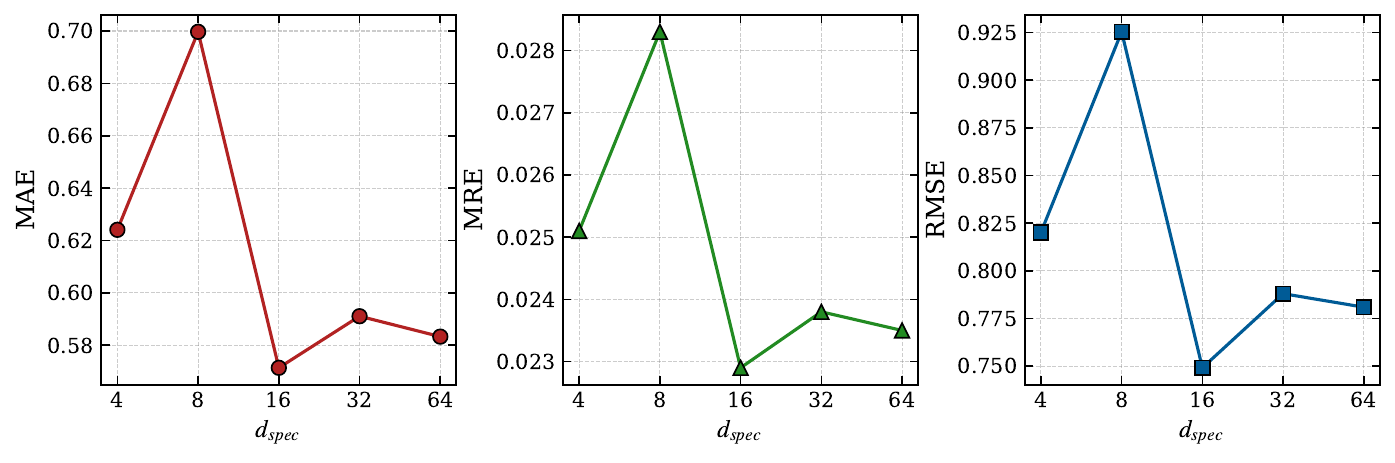}
    \caption{Performance of SG-NTF with different spectral dimensionalities $d_{spec}$ on the PCTemp (10:90) dataset.}
    \label{fig:sensitivity}
\end{figure}

Figure \ref{fig:sensitivity} illustrates the impact of the 
spectral dimensionality $d_{spec} \in \{4, 8, 16, 32, 64\}$ on 
SG-NTF using the PCTemp (10:90) dataset. The model 
attains optimal completion accuracy across all three metrics 
at $d_{spec} = 16$. A restricted dimensionality 
 yields higher estimation 
errors, indicating that insufficient spectral bases fail 
to accurately abstract complex continuous temporal periodicities. 
Conversely, expanding $d_{spec}$ beyond 16 causes slight 
performance degradation. This implies that excessive \hl{spectral bases} 
induce over-parameterization, exacerbating the overfitting risk under extreme 
data sparsity. Consequently, $d_{spec} = 16$ is consistently adopted to achieve an optimal 
balance between spectral representational capacity and model complexity.

\subsection{\hl{Gating Sparsity Analysis}}
\label{sec:sparsity}

\begin{figure}[htbp]
    \centering
    \includegraphics[width=0.85\linewidth]{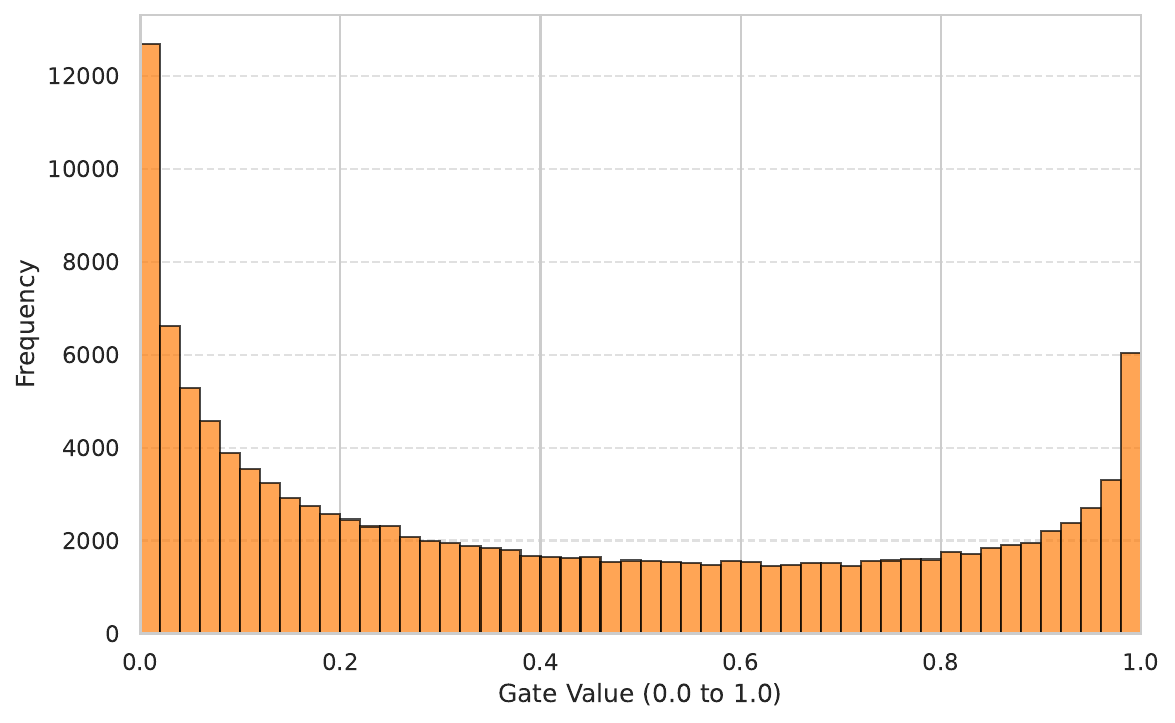}
    \caption{Distribution of gating activations $\mathbf{g}$}
    \label{fig:sparsity}
\end{figure}

Figure \ref{fig:sparsity} \hl{illustrates the empirical distribution 
of gate activations $\mathbf{g}$ over 1,000 sampled spatiotemporal 
contexts using the model attaining minimal 
MAE on the PCTemp (10:90) dataset. 
The distribution exhibits bimodal sparsity, characterized 
by a concentration of activations near 0. This suppression of latent 
interaction features validates the core zeroization mechanism, as the 
STCG module filters task-irrelevant components to refine the latent manifold.}

\section{Conclusion}
This paper proposes the SG-NTF model for accurate latent 
factorization of HDI tensors. By integrating the LFFM and STCG modules, it effectively parameterizes 
location-specific spatial responses to continuous temporal periodicities. This parameter-efficient formulation captures 
complex spatiotemporal dynamics, achieving a balance between representational capacity and architectural complexity 
for tensor completion.

\begin{table}
\centering
\caption{Ablation Study on PCTemp (10:90).}
\label{tab:ablation}
\setlength{\tabcolsep}{1.5mm}{
\begin{tabular}{lcc|ccc}
\toprule
\textbf{Model Variants} & \textbf{LFF} & \textbf{SC} & \textbf{MAE} & \textbf{MRE} & \textbf{RMSE} \\
\midrule
Base NeuTucF (M3) & $\times$ & $\times$ & 0.8575 & 0.0349 & 1.1429 \\
SG-NTF \textit{w/o} LFF & $\times$ & $\checkmark$ & 0.8534 & 0.0348 & 1.1397 \\ 
SG-NTF \textit{w/o} SC  & $\checkmark$ & $\times$ &  0.7163 &  0.0290&  0.9650\\ 
\midrule
\textbf{SG-NTF (Full Model)} & $\checkmark$ & $\checkmark$ & \textbf{0.5714} & \textbf{0.0229} & \textbf{0.7491} \\
\bottomrule
\multicolumn{6}{l}{\scriptsize \textbf{LFF}: Learnable Fourier Feature; \textbf{SC}: Spatial Context in Gating.}
\end{tabular}}
\end{table}

\vspace{12pt}
\end{document}